\documentclass[10pt,twocolumn,letterpaper]{article}

\usepackage{iccv}              

\definecolor{iccvblue}{rgb}{0.21,0.49,0.74}
\usepackage[pagebackref,breaklinks,colorlinks,allcolors=iccvblue]{hyperref}

\usepackage{wrapfig}
\usepackage{duckuments}
\usepackage{CJKutf8}
\usepackage{algorithm}  
\usepackage{algorithmicx}  
\usepackage{algpseudocode}  
\usepackage{overpic}

\definecolor{MyDarkBlue}{rgb}{0,0.5,1}
\definecolor{MyDarkGreen}{rgb}{0.02,0.6,0.02}
\definecolor{MyDarkRed}{rgb}{0.8,0.02,0.02}
\definecolor{MyDarkOrange}{rgb}{0.40,0.2,0.02}
\definecolor{MyYellow}{rgb}{1,0.55,0}
\definecolor{MyPurple}{RGB}{111,0,255}
\definecolor{MyRed}{rgb}{1.0,0.0,0.0}
\definecolor{MyGold}{rgb}{0.75,0.6,0.12}
\definecolor{MyDarkgray}{rgb}{0.66, 0.66, 0.66}
\definecolor{default}{RGB}{0,0,0}

\newcommand\bb[1]{\textbf{#1}}
\newcommand\ti[1]{\textit{#1}}

\renewcommand\etal{\textit{et al.}} 
\renewcommand\eg{\textit{e.g., }}
\renewcommand\ie{\textit{i.e., }}

\newcommand{\model}{DisWM} 
\renewcommand{\eqref}[1]{Eq.~(\ref{#1})} %
\newcommand{\figref}[1]{Figure~\ref{#1}} %
\newcommand{\tabref}[1]{Table~\ref{#1}} %
\newcommand{\appref}[1]{\underline{Supplementary Material~\ref{#1}}} %
\def\paperID{2719} 
\def\confName{ICCV}
\def\confYear{2025}

\title{Disentangled World Models: Learning to Transfer Semantic \\ Knowledge from Distracting Videos for Reinforcement Learning}

\author{
Qi Wang$^{1,2,3}$\footnotemark[1]\thanks{Equal contribution.} \
Zhipeng Zhang$^{4,5}$\footnotemark[1] \
Baao Xie$^{2,3}$\footnotemark[1] \
Xin Jin$^{2,3}$\thanks{Corresponding author:  Xin~Jin~\textless jinxin@eitech.edu.cn\textgreater.} \ 
Yunbo Wang$^{1}$ \\
Shiyu Wang$^{5,6}$ \ 
Liaomo Zheng$^{5,6}$ \
Xiaokang Yang$^{1}$ \
Wenjun Zeng$^{2,3}$  \\
$^1$ MoE Key Lab of Artificial Intelligence, AI Institute, Shanghai Jiao Tong University \\
$^2$ Ningbo Institute of Digital Twin, Eastern Institute of Technology, Ningbo, China\\
$^3$ Ningbo Key Laboratory of Spatial Intelligence and Digital Derivative, Ningbo, China\\
$^4$ University of Chinese Academy of Sciences \\
$^5$ Shenyang Institute of Computing Technology, Chinese Academy of Sciences\\
$^6$ Shenyang CASNC Technology Co., Ltd\\
\textcolor{magenta}{\url{https://qiwang067.github.io/diswm}}
}

\begin{document}
\maketitle
\begin{abstract}
Training visual reinforcement learning (RL) in practical scenarios presents a significant challenge, \textit{i.e.,} RL agents suffer from low sample efficiency in environments with variations.
While various approaches have attempted to alleviate this issue by disentangled representation learning, these methods usually start learning from scratch without prior knowledge of the world.
This paper, in contrast, tries to learn and understand underlying semantic variations from distracting videos via offline-to-online latent distillation and flexible disentanglement constraints.
To enable effective cross-domain semantic knowledge transfer, we introduce an interpretable model-based RL framework, dubbed Disentangled World Models (DisWM).
Specifically, we pretrain the action-free video prediction model offline with disentanglement regularization to extract semantic knowledge from distracting videos.
The disentanglement capability of the pretrained model is then transferred to the world model through latent distillation.
For finetuning in the online environment, we exploit the knowledge from the pretrained model and introduce a disentanglement constraint to the world model.
During the adaptation phase, the incorporation of actions and rewards from online environment interactions enriches the diversity of the data, which in turn strengthens the disentangled representation learning.
Experimental results validate the superiority of our approach on various benchmarks.
\end{abstract}
\section{Introduction}
Visual reinforcement learning (VRL) presents a promising approach for training agents within complex environments~\cite{laskin2020curl,pan2022iso,hafner2022deep,zhang2024prelar,li2025open}. 
However, VRL frequently suffers from performance degradation in practical scenarios due to the complexity, volatility, and visual distractions in environments. Even minor environmental variations can result in significant pixel-level shifts, making the trained VRL policies ineffective or suboptimal~\cite{dunion2023temporal,dunion2023conditional}. For instance, a slight change in lighting conditions can affect an object's appearance (\eg color, shadow, or other visual attributes). Therefore, it is crucial to enhance models with interpretability, enabling them to perceive, learn, and understand the semantic environmental variations.

\begin{figure}[t]
    \centering
    \includegraphics[width=\linewidth]{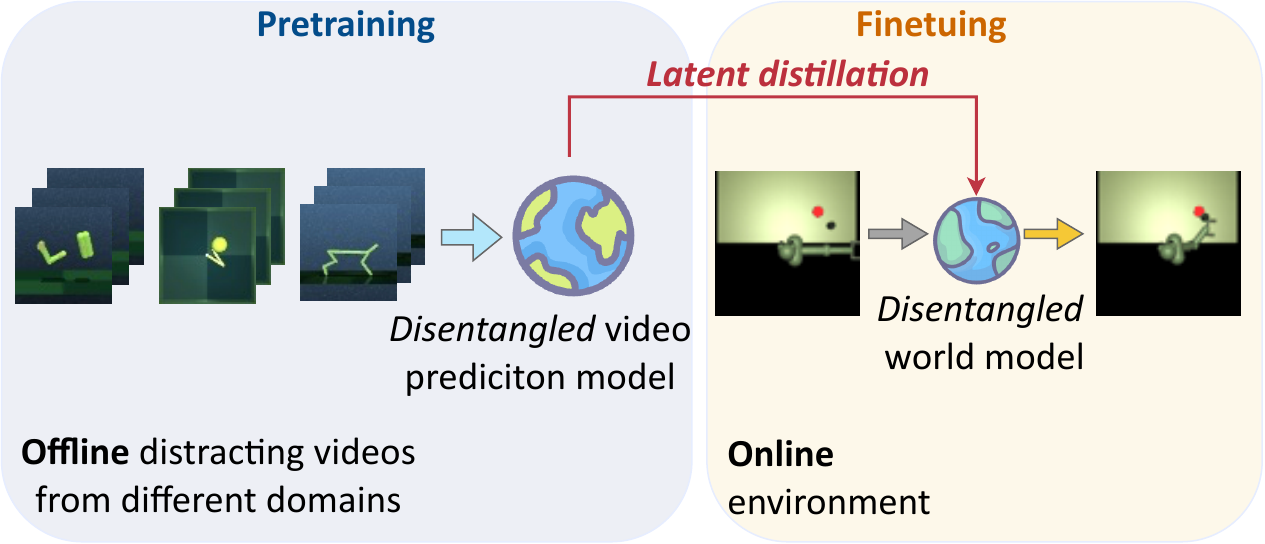}
    \caption{Overview of our proposed framework. The key idea is to leverage distracting videos for semantic knowledge transfer, enabling the downstream agent to improve sample efficiency on unseen tasks.
     }
    \label{fig:teaser_fig}
    \vspace{-15pt}
\end{figure}

\begin{table*}[t]
    \caption{MuJoCo (\textit{downstream domain}) vs. DMC (\textit{accessible distracting videos}).} 
    \label{tab:dmc_pusher_cmp}
    \vspace{-10pt}
    \setlength\tabcolsep{5pt}
    \begin{center}
    \footnotesize
    \centering
    \begin{tabular}{l|ccc}
    \toprule
    & Video: \textit{DMC} & Target: \textit{MuJoCo} &  Similarity / Difference \\
    \midrule
    Task& Reacher Easy & Pusher & Relevant robotic control tasks\\
    Dynamics&  Two-link planar & Multi-jointed robot arm  & Different \\
    Action space & Box(-1, 1, (2,), float32) & Box(-2, 2, (7,), float32) & Different\\
    Reward range& [0, 1]  & [-4.49, 0] &  Different \\  
    \bottomrule
    \end{tabular}
    \end{center}
    \vspace{-10pt}
    \end{table*}

Disentangled representation learning (DRL) presents a promising approach to addressing the interpretability challenges inherent in the ``black-box'' nature of deep learning algorithms. Fundamentally, DRL approaches mimic the cognitive processes of biological intelligence, where understanding the world is facilitated by decomposing observations into distinct and independent factors ~\cite{bengio2013representation,higgins2017beta,xie2023navinerf,wang2024disentangled,xie2024graph}. 
In this form, when a factor of variation is changed~(\eg color), only a small portion of features in the disentangled representation will be affected, enabling the agent to recover performance quickly.
Several studies have explored the integration of DRL algorithms in the domain of VRL. 
For example, Higgins \etal~\cite{higgins2017darla} trained a $\beta$-VAE offline to obtain disentangled representations for reinforcement learning.
TED~\cite{dunion2023temporal} adopts a self-supervised auxiliary task to learn temporally disentangled representations for reinforcement learning. 
Additionally, Dunion \etal~\cite{dunion2023conditional} introduced conditional mutual information to achieve a disentangled representation with correlated data.
However, existing methods typically learn the representations from scratch, lacking any prior knowledge of the world.
These approaches often require extensive interactions with the environment to acquire desired behaviors.

Towards this challenge, we introduce a model-based interpretable VRL framework, dubbed \textbf{Dis}entangled \textbf{W}orld \textbf{M}odels~(\textbf{DisWM}), which leverages prior knowledge extracted from distracting videos to facilitate the learning of unseen downstream tasks through latent distillation. 
It is crucial to note that \bb{distracting videos refer to videos with visual distractions}, which are beneficial for learning disentangled representations.
Specifically, as depicted in \figref{fig:teaser_fig}, our framework consists of two phases: first, we pretrain a DRL encoder to learn disentangled latent representations from distracting videos. 
By doing so, the pretrained DRL encoder is ``knowledgeable'' in terms of representation disentanglement.
Subsequently, we finetune an orthogonally designed world model with dual constraints of disentanglement and distillation, leveraging semantic knowledge transferred from the pretrained model via offline-to-online latent distillation.
Another benefit of disentangled world model adaptation is that incorporating actions and rewards from online interactions with the environment enriches the diversity of the visual observations, which in turn strengthens the process of disentangled representation learning. 
It is worth mentioning that, as a cross-domain framework, \model{} does not require the pretraining videos to originate from the same domain as the downstream tasks. 

Experimental results demonstrate the effectiveness of our proposed approach in improving the sample efficiency of VRL agents across our modified DeepMind Control and MuJoCo \textit{Pusher}. The contributions of this work can be summarized as follows:
\begin{itemize}
    \item  We frame the problem of learning interpretable VRL agents as a domain transfer learning problem. The key idea is to extract semantic knowledge from distracting videos and transfer this disentanglement capability to downstream control tasks.
    \item  We present \model{}, an approach that follows the pretraining-finetuning paradigm using distracting videos, incorporating specific techniques of offline-to-online latent distillation and flexible disentanglement constraints.
\end{itemize}

\section{Problem Setup}
\label{sec:problem_setup}

We formulate visual reinforcement learning as a partially observable Markov decision process (POMDP) that uses DMC and MuJoCo \textit{Pusher} as the test bench.
Specifically, we concentrate on scenarios where videos without actions and rewards are accessible, enabling world knowledge transfer.
The goal is to maximize the cumulative reward of the target POMDP $\left\langle\mathcal{O}, \mathcal{A}, \mathcal{T}, \mathcal{R}, \gamma \right\rangle$ by transferring the shared world knowledge from the videos. 
These notations correspond to the visual observation space, the action space, the transition probabilities, the reward function, and the discount factor, respectively.

For instance, in one of the cross-domain experiments, we use MuJoCo as the downstream domain and the frames collected from DMC as the distracting video.
Table \ref{tab:dmc_pusher_cmp} highlights the differences between the two domains in terms of visual appearances, physical dynamics, action spaces, and reward functions.

\section{Method}

\begin{figure*}[htb]
    \centering
    \includegraphics[width=\linewidth]{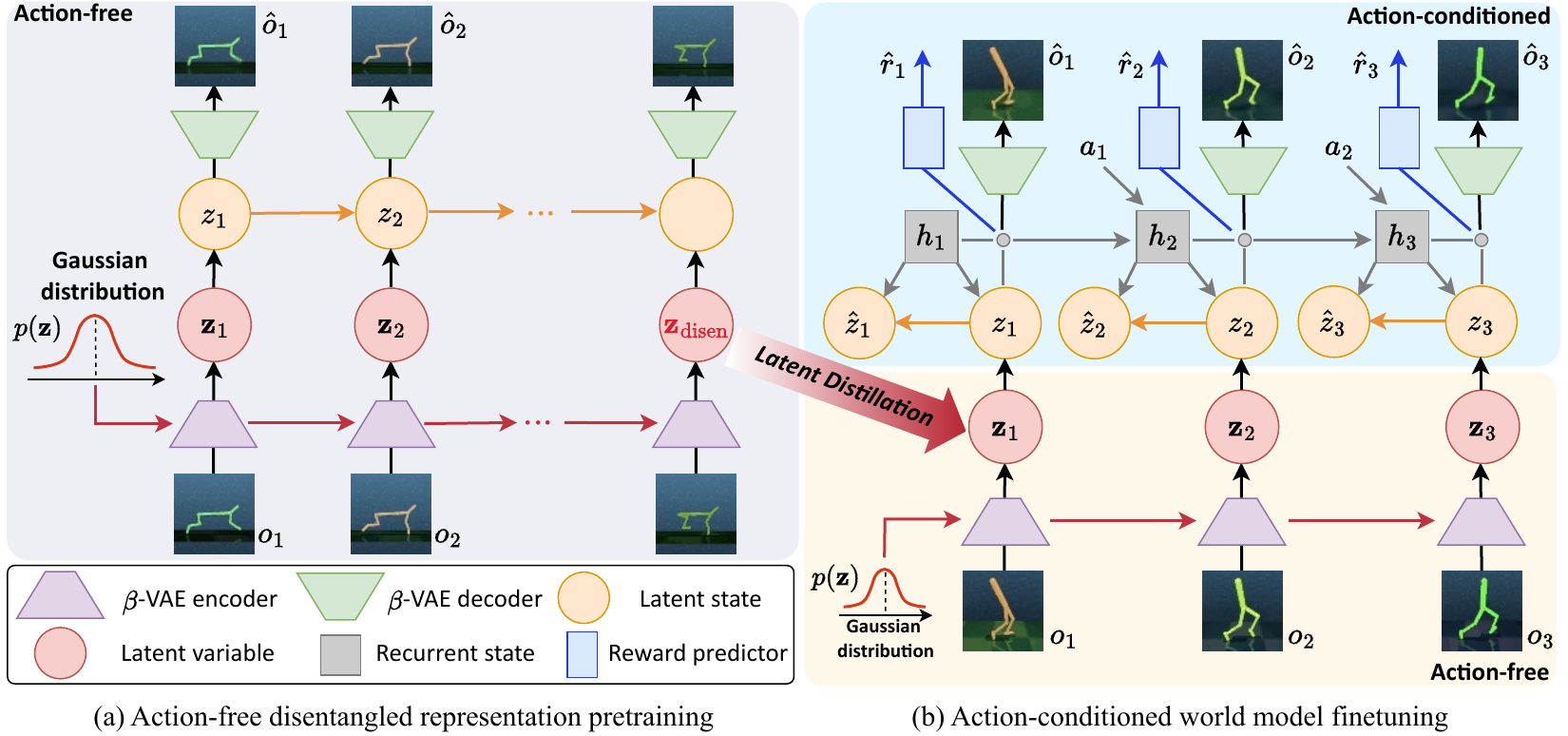}
    \caption{Architecture of Disentangled World Models. \textbf{(a)}~The action-free video prediction model with disentanglement constraints is pretrained on distracting videos offline for the well-disentangled latent variable $\bb{z}_{\text{disen}}$, which extracts semantic knowledge from the visual observations. 
    The disentangled capability of $\bb{z}_{\text{disen}}$ is then transferred to the world model through latent distillation.
    \textbf{(b)}~The action-conditioned world model is finetuned through disentanglement regularization online, which encourages the factorized representation.
    Moreover, the incorporation of actions and rewards enriches the diversity of the data, which in turn strengthens the disentangled representation learning. 
    }
    \label{fig:pipeline}
\end{figure*}

\subsection{Overview of \model{}}
In this section, we present the details of \model{}, which involves three main stages~(see \figref{fig:pipeline}):
\begin{enumerate}
    \renewcommand{\labelenumi}{\alph{enumi})}
        \item \textit{Disentangled representation pretraining}: Pretrain a DRL-based video prediction model from distracting videos to extract disentangled features.
        \item \textit{Offline-to-online latent distillation}: Transfer the semantic knowledge from the pretrained model to the world model via cross-domain latent distillation.
        \item \textit{Disentangled world model adaptation}: 
        Finetune the downstream agent with disentanglement constraints by incorporating the action and reward information.
    \end{enumerate}

\subsection{Disentangled Representation Pretraining}
To extract well-disentangled representations that can be transferred to the downstream world model, we first train a video prediction model on distracting videos without incorporating action information~(\textbf{Lines 4-8} of Alg. \ref{algo:overall}). This model comprises three key components: 
(i) the posterior learner that encodes the observation $o_t$ to latent state $z_t$ via $\beta$-VAE encoder\footnote{$\beta$-VAE~\cite{higgins2017beta} is a variant of autoencoder that introduces a hyperparameter $\beta$ to balance the reconstruction quality and disentanglement capability.}, which serves as a typical DRL framework to extract latent features $\mathbf{z}_t$ from observations,
(ii) the prior module that predicts future latent states based on historical states, without directly relying on the current observation $o_t$,
and 
(iii) the $\beta$-VAE-based decoder that reconstructs $\hat{o}_t$ from the latent state $z_t$. 
Concretely, the model can be formulated as follows:
\begin{equation}
    \label{eq:pretrain_model}
    \begin{alignedat}{3}
    &\text{$\beta$-VAE encoder: }  &{\mathbf{z}}_t & = e_{\phi^{\prime}}(o_t)\\
    &\text{Posterior state: }  &{z}_t & \sim q_{\phi^{\prime}}(z_t \mid z_{t-1}, \mathbf{z}_t)\\
    &\text{Prior state: }  &\hat{z}_t & \sim p_{\phi^{\prime}}(\hat{z}_t \mid z_{t-1})\\
    &\text{Reconstruction: } &\hat{o}_t & \sim p_{\phi^{\prime}}(\hat{o}_t \mid z_t)\\
    &\text{Isotropic unit Gaussian: } &p(\mathbf{z}) &=\mathcal{N}(\mathbf{0}, I).
    \end{alignedat}
\end{equation}
where $\phi^{\prime}$ denotes the parameters of the model.
The $\beta$-VAE-based video prediction model is trained to minimize the following loss function:
\begin{equation}
    \small
    \begin{aligned}
    \label{eq:pretrain_loss}
    \mathcal{L}(\phi^{\prime}) = \ & \mathbb{E}_{q_{\phi^{\prime}}}
    \Big[
    \sum_{t=1}^T \underbrace{-\ln p_{\phi^{\prime}}(o_t \mid z_t)}_{\text {image reconstruction}} 
    \\&+\underbrace{\beta_1 \mathrm{KL}\left[q_{\phi^{\prime}}\left(z_t \mid z_{t-1}, o_t\right) \| p_{\phi^{\prime}}\left(\hat{z}_t \mid z_{t-1}\right)\right]}_{\text {action-free KL loss }}
    \\&\underbrace{+ \beta_2 \mathrm{KL}\left[q_{\phi^{\prime}}(\mathbf{z}_t \mid o_t) \ \| \ p(\mathbf{z}_t)\right]}_{\text{disentanglement loss}} \Big].
    \end{aligned}    
\end{equation}
The variantional posterior distribution $q_{\phi^{\prime}}(\mathbf{z}_t \mid o_t)$ is encouraged to be close to the standard multivariate Gaussian distribution $\mathcal{N}(\mathbf{0}, I)$ to strengthen the orthogonality and disentanglement of the latent space. The importance of the disentanglement loss term is governed by $\beta_2$.

\subsection{Offline-to-Online Latent Distillation}
After the offline pretraining with the distracting videos, the model is fine-tuned online to adapt to the downstream task by integrating actions and rewards~(\textbf{Lines 13} of Alg. \ref{algo:overall}). 
A straightforward approach to transfer disentangled features involves initializing the action-conditioned world model with checkpoints obtained from the pretrained video prediction model. 
Nevertheless, it may experience a potential mismatch issue caused by the discrepancies between the two domains in visual appearances and physical dynamics.
Directly applying the pretraining-finetuning paradigm for downstream tasks tends to overwrite the disentangled information encoded in the pretrained latent features, leading to decreased performance when there are large domain discrepancies between the source and the target domains.

Through comprehensive pretraining on distracting videos that contain diverse visual variations, the video prediction model thus builds an interpretable and orthogonality latent space. In this space, the latent variable $\mathbf{z}_{\text{disen}}$ achieves a high degree of disentanglement. To exploit the prior semantic knowledge from the pretrained model and improve the sample efficiency of the downstream tasks, we introduce an offline-to-online latent distillation. This approach enables the disentangling capability of $\mathbf{z}_{\text{disen}}$ from the pretrained model to be effectively transferred to the latent variable $\mathbf{z}_{\text{task}}$ of the world model. Specifically, this is achieved by minimizing the Kullback-Leibler (KL) divergence between the latent distributions of the two domains. The corresponding distillation loss $\mathcal{L}_{\text{distill}}$ can be formulated as follows:
\begin{equation}
    \label{eq:distillation_loss}
\mathcal{L}_{\text{distill}}=\text{KL}\left(\mathbf{z}_{\text{disen}} \| \mathbf{z}_{\text {task }}\right)=\sum \mathbf{z}_{\text {disen}} \cdot \log \left(\frac{\mathbf{z}_{\text {disen }}}{\mathbf{z}_{\text {task }}}\right)
\end{equation}

\subsection{Disentangled World Model Adaptation}
By obtaining well-disentangled representations of $\mathbf{z}_\text{disen}$ and employing the latent distillation for knowledge transfer, we then propose a DRL-based world model $\mathcal{M}_{\phi}$, designed to harness these features to enhance interoperability and robustness against environmental variations~(\textbf{Lines 14-15} of Alg. \ref{algo:overall}). The components of $\mathcal{M}_{\phi}$ can be detailed as follows:

\vspace{-9pt}
\begin{equation}
\label{eq:wm}
\begin{alignedat}{3}
&\text{Recurrent transition: } & h_t &= f_{\phi}(h_{t-1}, z_{t-1}, a_{t-1})  \\
&\text{$\beta$-VAE encoder: } & {\mathbf{z}}_t & \sim e_{\phi}(o_t)\\
&\text{Posterior state: } & {z}_t & \sim q_{\phi}(z_t \mid h_{t},\mathbf{z}_t)\\
&\text{Prior state: } & {\hat{z}}_t & \sim p_{\phi}(\hat{z}_t \mid h_{t})\\
&\text{Reconstruction: } & \hat{o}_t &\sim p_{\phi} (\hat{o}_t\mid h_t,z_t) \\
&\text{Reward prediction: } & \hat{r}_t &\sim r_{\phi}(\hat{r}_t \mid h_t,z_t) \\
&\text{Discount factor: } &\hat{\gamma}_t &\sim p_{\phi}(\hat{\gamma}_t \mid h_t,z_t) \\
&\text{Isotropic unit Gaussian: } &p(\mathbf{\mathbf{z}}) &=\mathcal{N}(\mathbf{0}, I),
\end{alignedat}
\end{equation}
where $\phi$  represents the combined parameters of the world model. 
We train  $\mathcal{M}_{\phi}$ on the sampled data from the replay buffer $\mathcal{B}$ with the following loss function:

\begin{equation}
    \small
    \begin{aligned}
    \label{eq:beta-vae_based_world_model_loss}
    \mathcal{L}(\phi) = \ & \mathbb{E}_{q_{\phi}}
    \Big[
    \sum_{t=1}^T \underbrace{-\ln p_{\phi}(o_t \mid h_t, z_t)}_{\text {image reconstruction}} \underbrace{-\ln r_{\phi}(r_t \mid h_t, z_t)}_{\text {reward prediction}} \\ &\underbrace{-\ln p_{\phi}(\gamma_t \mid h_t, z_t)}_{\text {discount prediction }}  \underbrace{+\alpha \ \mathrm{KL}\left[q_{\phi}(z_t \mid h_t, o_t) \ \| \ p_{\phi}(\hat{z}_t \mid h_t)\right]}_{\text{KL divergence}}
    \\
    &\underbrace{+ \beta \mathrm{KL}\left[q_{\phi}(\mathbf{z}_t \mid o_t) \ \| \ p(\mathbf{z}_t)\right]}_{\text{disentanglement loss}} 
    + \underbrace{\eta \mathcal{L}_{\text{distill}}}_{\text{distillation loss}}
     \Big].
    \end{aligned}    
    \end{equation}
where $\beta$ is a hyperparameter used to balance reconstruction quality and disentanglement capability.
In this adaptation stage, $\eta$ serves as a hyperparameter that gradually decreases from 0.1 to 0.01.
Intuitively, $\eta$ controls the progressive adaptation of the world model with the shared world knowledge transfer from the pretrained video prediction model.  

Through the comprehensive training processes of this framework, we equip the DisWM with the capability to learn and understand the underlying semantic representations. 
This enhancement enables the model to be less sensitive to environmental variations, such as changes in object colors, positions, and backgrounds. 
Furthermore, by incorporating actions and rewards during the finetuning phase, the world model can generate data with more diverse representations, thereby improving the disentangled representation learning.
For behavior learning, we utilize the actor-critic method that is in line with DreamerV2~\citep{hafner2021mastering}~(\textbf{Lines 16-18} of Alg. \ref{algo:overall}). For more details of behavior learning, please refer to \appref{sec:bl}.

\begin{algorithm*}[t]
    \caption{The training pipeline of \model{}.}
    \label{algo:overall}
    \begin{algorithmic}[1]
    \small
    \State \textbf{Hyperparameters: }{$H$: Horizon of latent imagination.} 
    \State \textbf{Require: }{Distracting video dataset $\mathcal{D}$.} 
    \State \textbf{Initialize:} Parameters of the model $\{\phi, \psi, \xi\}$. 
    \For{training  step $t=1,2,\ldots,K_1$}
    \Comment{Disentangled representation pretraining}
    \State Sample random minibatch $\{o_{t}\}_{t=1}^{T} \sim \mathcal{D}.$
    \State Obtain Gaussian prior $\mathbf{z}_t$ from Isotropic unit Gaussian $\mathcal{N}(\mathbf{0}, I)$.
    \State Pretrain the action-free video prediction model with disentanglement regularization by minimizing \eqref{eq:pretrain_loss}.  
    \EndFor 
    \State Train the random agent and collect a replay buffer $\mathcal{B}$.
    \While{not converged}
    \For{training step $t=1,2,\ldots,K_2$}
         \State Sample $\{(o_{t}, a_{t}, r_{t})\}_{t=1}^{T} \sim \mathcal{B}$. 
         \State Distill the disentangled features to the world model using \eqref{eq:distillation_loss}.
         \Comment{Offline-to-online latent distillation}
         \State Obtain Gaussian prior $\mathbf{z}_t$ from Isotropic unit Gaussian $\mathcal{N}(\mathbf{0}, I)$. \Comment{Disentangled world model adaptation}
        \State Train the world model $\mathcal{M}_\phi$ with latent distillation and disentanglement constraints  using \eqref{eq:beta-vae_based_world_model_loss}. 
        \State Generate $\{(\hat{z}_i, \hat{a}_i)\}_{i=t}^{t+H}$ using $\pi_\psi$ and $\mathcal{M}_\phi$. \Comment{Behavior learning} 
        \State Train the critic $v_\xi$ over $\{(\hat{z}_i, \hat{a}_i)\}_{i=t}^{t+H}$. 
        \State Train the actor $\pi_\psi$ over $\{(\hat{z}_i, \hat{a}_i)\}_{i=t}^{t+H}$. 
    \EndFor
    \State $o_1$ $\leftarrow$ \texttt{env.reset()}
    \Comment{Environment interaction}
    \For{time step $t=1,2,\ldots,T$}
        \State Sample $\hat{a}_t$ $\sim$ $\pi_{\psi}(\hat{a}_t \mid \hat{z}_t)$.
        \State $r_t, o_{t+1} \leftarrow$ \texttt{env.step}($\hat{a}_t$).
    \EndFor
    \State Append data to the replay buffer $\mathcal{B}$. 
    \EndWhile
  \end{algorithmic}
  \end{algorithm*}
\section{Experiments}
\subsection{Experimental Setups} 
\paragraph{Benchmark} 
We evaluate \model{} on DeepMind Control Suite~(DMC)~\cite{tassa2018deepmind}, MuJoCo \textit{Pusher}~\cite{todorov2012mujoco}, and DrawerWorld~\cite{wang2021unsupervised}.
DMC is a widely adopted benchmark with comprehensive and flexible robotic-control tasks. 
For DMC benchmark, we use 5 tasks, \ie \textit{Walker Walk}, \textit{Cheetah Run}, \textit{Hopper Stand}, \textit{Finger Spin}, \textit{Cartpole Swingup}. 
In MuJoCo \ti{Pusher}, a multi-jointed robotic arm is employed to manipulate a target cylinder~(object). 
The goal is to move the object to a designated target position using the robot's end effector~(fingertip). 
The agent receives the negative reward, which is a combination of three components: the distance between the fingertip and the target, the distance between the object and the target goal position, and a penalty for large actions.
DrawerWorld is a modified Metaworld~\cite{yu2019meta} benchmark designed to evaluate texture adaptability in manipulation tasks. It includes five additional textures from realistic photos and \textit{grid} texture.
During training, we initially employ the \textit{grid} texture and then change to the \textit{wood} texture midway, while adopting the \textit{metal} texture exclusively for evaluation. The corresponding results on DrawerWorld are reported in \appref{sec:results_drawerworld}. 
Furthermore, the introduction of compared baselines is given in \appref{sec:compared_baselines}. 

%

\begin{figure*}[htb]
    \centering
        \includegraphics[width=0.8\linewidth]{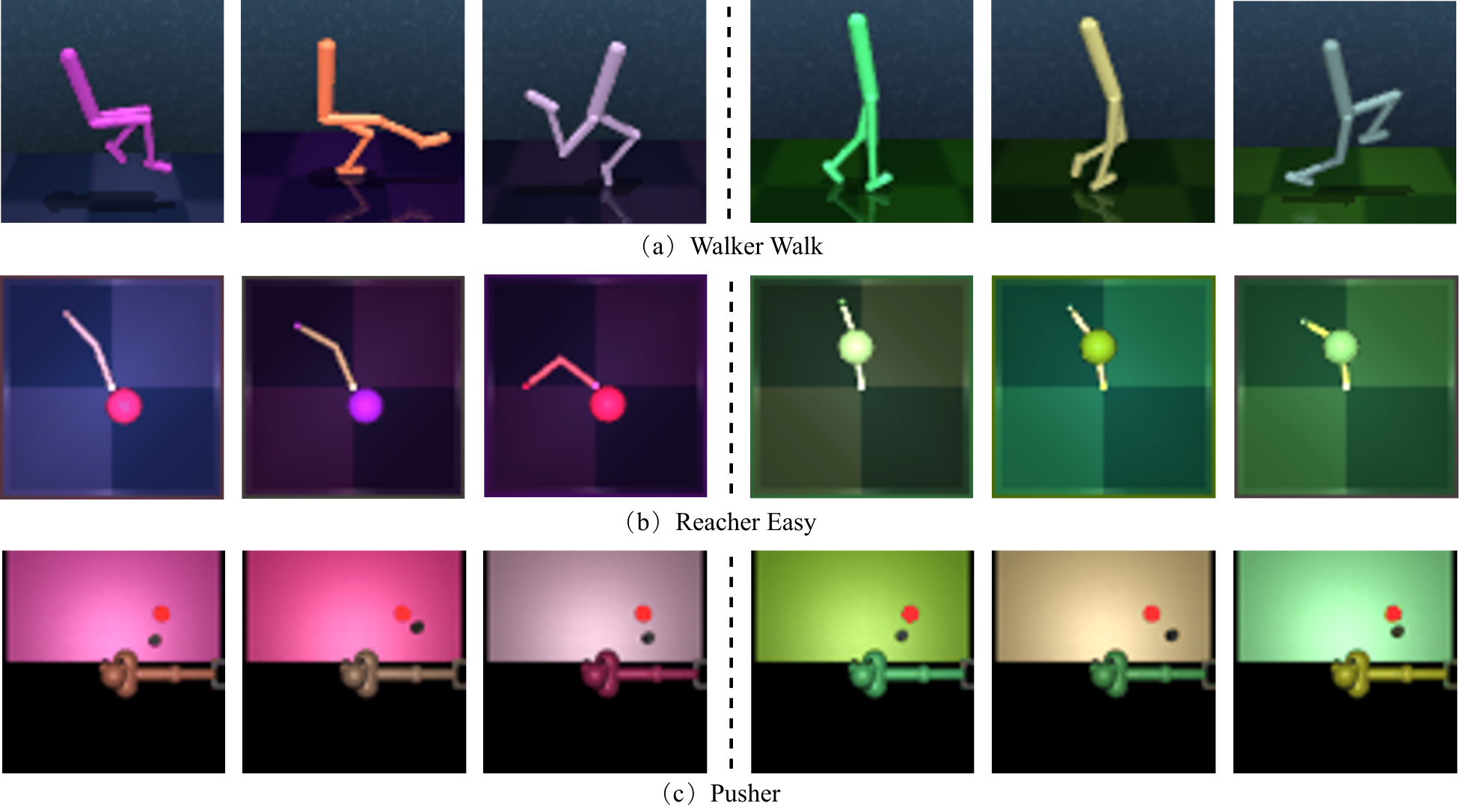}
    \caption{Example image observations of our modified DMC and MuJoCo \textit{Pusher} with color distractors.}
    \label{fig:showcase}
\end{figure*}

\paragraph{Implementation  details.} 
The visual observations of online finetuning stages are resized to $64 \times 64$ pixels. 
Inspired by APV~\cite{seo2022reinforcement}, we build distracting video datasets with 1M frames using DreamerV2~\cite{hafner2021mastering} to interact with the environments with visual color distractors. This video datasets consist of the samples stored in the replay buffer throughout the training process until the agent achieves maximum score.
For tasks in the DMC benchmark, the training steps of the agent are limited to $1\times 10^6$ environment steps. Each run of \model{} requires roughly $5$GB of VRAM and takes around $16$ hours to train on a single RTX 3090 GPU. 
The dimensions of well-disentangled latent $\mathbf{z}_{\text{disen}}$ and downstream task latent $\mathbf{z_{\text{task}}}$ are both set to 20 in our approach. 
In \figref{fig:showcase}, we showcase the example observations of various tasks with color distractors. 
We train the agent using a fixed set of colors, where the RGB values are varied within a restricted range around the original values.
Furthermore, at the midpoint of the training process, we change to a different color scheme for varying distractors.

\begin{figure*}[thb]
    \centering
    \begin{overpic}[width=\linewidth]{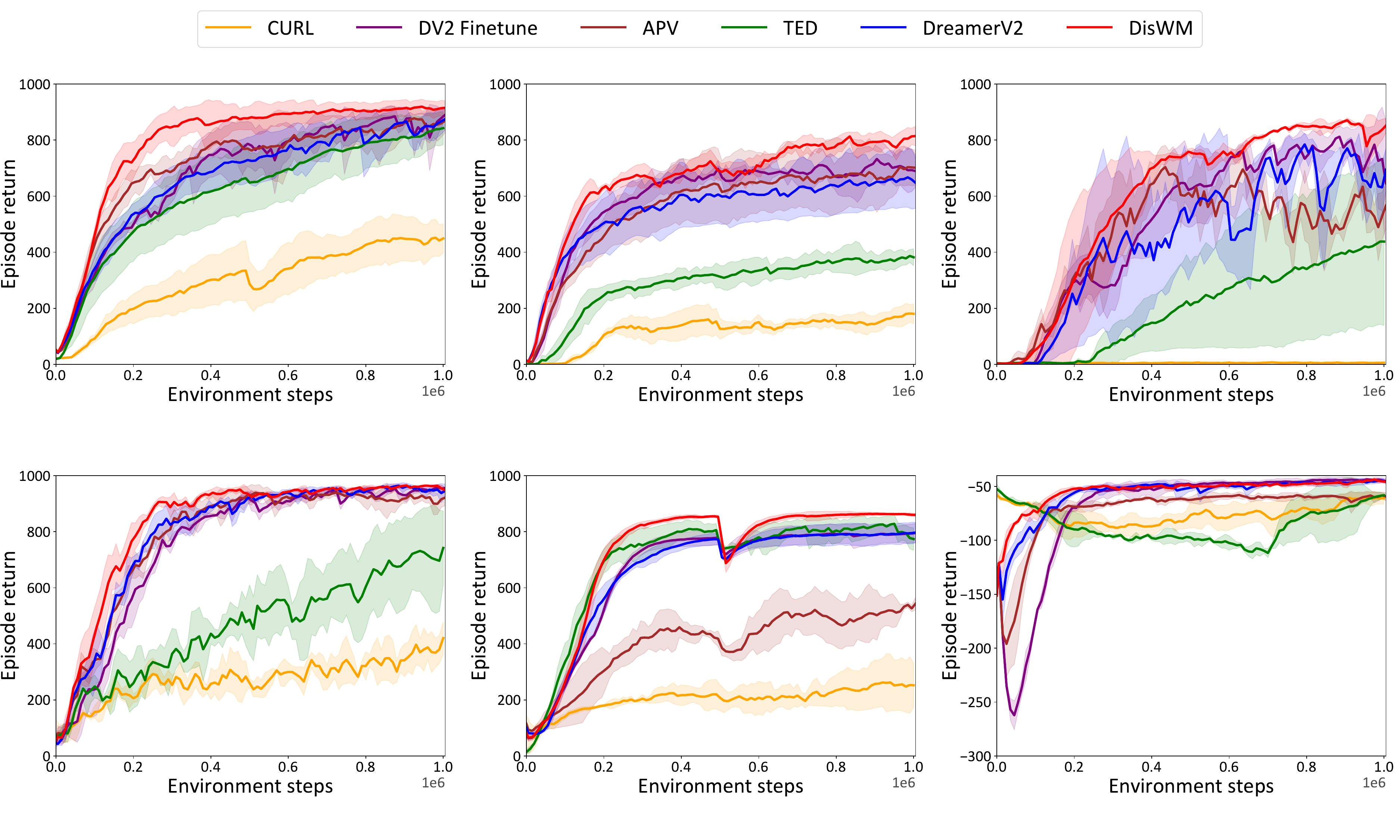}
    \put(8,29){\footnotesize (a) \ti{Cheetah Run} $\rightarrow$ \ti{Walker Walk}}
    \put(40.5,29){\footnotesize (b) \textit{Reacher Easy} $\rightarrow$ \textit{Cheetah Run}}
    \put(74,29){\footnotesize (c) \textit{Cheetah Run} $\rightarrow$ \textit{Hopper Stand}}
    \put(7,1){\footnotesize (d) \textit{Finger Spin} $\rightarrow$ \textit{Reacher Easy}}
    \put(40,1){\footnotesize (e) \textit{Finger Spin} $\rightarrow$ \textit{Cartpole Swingup}}
    \put(76,1){\footnotesize (f) \textit{Reacher Easy} $\rightarrow$ \textit{Pusher}}
    \end{overpic}
    \caption{Comparison of \model{} against visual RL baselines, including \ti{DreamerV2}~\cite{hafner2021mastering}, \ti{APV}~\cite{seo2022reinforcement}, \ti{DV2 Finetune}, \ti{TED}~\citep{dunion2023temporal}, \ti{CURL}~\citep{laskin2020curl}.
    }
    \label{fig:baseline_compare}
\end{figure*}

\begin{figure*}[htb]
    \centering
    \includegraphics[width=0.8\linewidth]{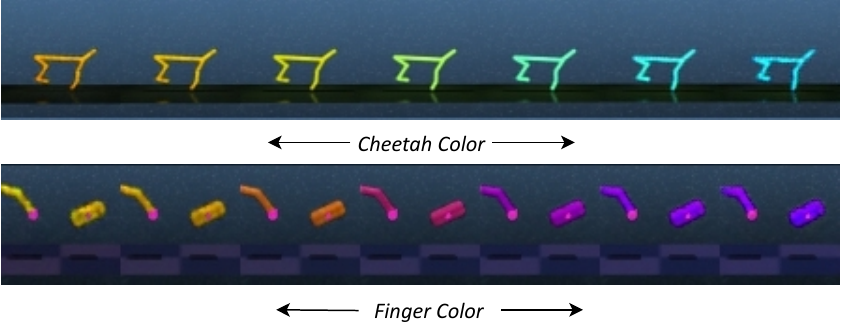}
    \vspace{-5pt}
    \caption{Visualization of traversals of $\beta$-VAE during the pretraining phase.}
    \label{fig:pretrain_dis}
\end{figure*}

\begin{figure*}[htb]
    \centering
    \includegraphics[width=0.8\linewidth]{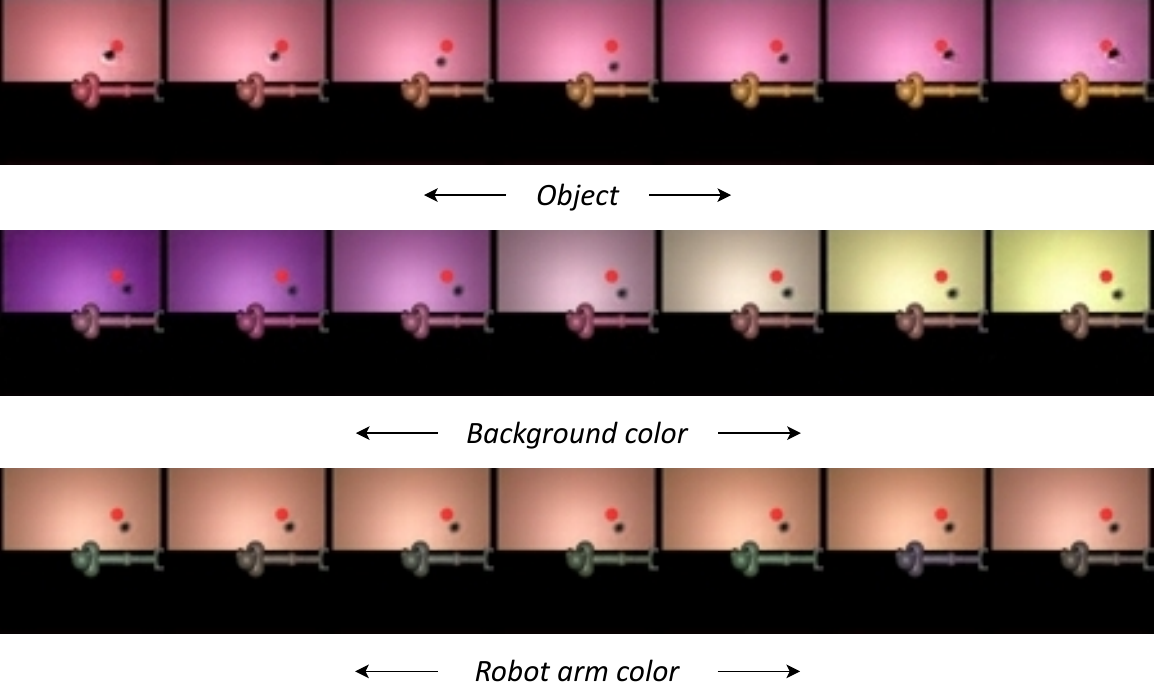}
    \caption{Fine-grained disentanglement results on MuJoCo \textit{Pusher} during world model adaptation phase.
    Each row in \ti{Pusher} images displays the traversal results on a specific attribute.} 
    \label{fig:finetune_dis}
\end{figure*}

\subsection{Main Comparison} 
We evaluate the sample efficiency and task performance for all the methods with the training curves of episode return. 
\figref{fig:baseline_compare} shows the performance of \model{} and all the baselines. 
Remarkably, it achieves better performance than TED~\cite{dunion2023temporal}, a method on top of RAD~\cite{laskin2020reinforcement} and tailored for environments with distractors. 
For the offline-to-online finetuning models, \textit{DV2 Finetune} achieves the second-best performance by transferring knowledge from the distracting videos. However, we observe a significant decline in sample efficiency, particularly in scenarios with large data distribution shifts between the source and target domains~(\eg DMC $\rightarrow$ MuJoCo). These shifts can occur in various aspects, including visual observation, physical dynamics, reward definition, or the action space of the robots.
Another crucial baseline is APV~\cite{seo2022reinforcement}, which focuses on transferring knowledge obtained from videos with a stacked latent prediction model.
Nevertheless, without environment-specific designs for visual distractors, directly training may eventually result in a decrease in performance in the downstream tasks.
The CURL model struggles to learn effective behavior policies, especially in \ti{Hopper Stand}.
Additional results on the challenging DMC \textit{Humanoid Walk} can be found in \appref{sec:res_dmc}.

Additionally, we present qualitative results in \figref{fig:pretrain_dis} and \figref{fig:finetune_dis}. 
\figref{fig:pretrain_dis} shows the traversals of $\beta$-VAE during the pretraining phase.
In each row of traversals, a distinct attribute varies, while other attributes remain constant, indicating that the pretrained model has successfully disentangled and learned this attribute, thereby improving the sample efficiency of the RL agent.
\figref{fig:finetune_dis} displays the fine-grained disentanglement results on MuJoCo \textit{Pusher} during the finetuning phase, demonstrating that the world model can effectively disentangle the variations.

\subsection{Model Analyses}
\paragraph{Ablation studies.}

\begin{figure*}[htb]
    \centering
    \includegraphics[width=\linewidth]{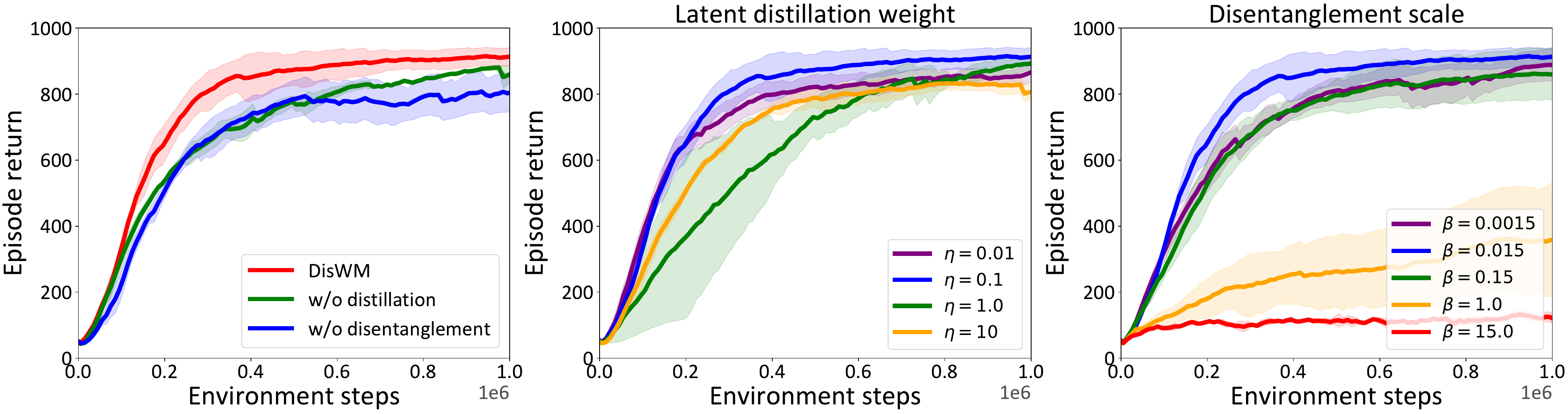}
    \caption{These figures illustrate the ablation studies and sensitivity analyses of \model{} on DMC \textit{Walker Walk} $\rightarrow$ \textit{Cheetah Run}.
    \bb{Left:} Comparison with \model{} without cross-domain latent distillation or disentanglement constraints.
    \bb{Middle:} The sensitivity analyses of cross-domain latent distillation weight. 
    \bb{Right:} The performance of \model{} with different disentanglement scale. 
    }
    \label{fig:ablation_sensitivity_results}
\end{figure*}

We conduct ablation studies to validate the effect of the latent distillation and disentanglement constraints.
\figref{fig:ablation_sensitivity_results}~ (Left) shows corresponding results in the DMC \textit{Walker Walk} $\rightarrow$ \textit{Cheetah Run}.  
The green curve shows that removing the latent distillation of \model{} results in a decreased performance, which indicates that the latent distillation is essential during the early training stage. 
For the model represented by the blue curve, we do not adopt disentanglement constraints for both pretrain and finetune stages. It can be seen that the necessity of introducing DRL-based training and disentangled representation significantly improves the learning efficiency of the agent.

\paragraph{Sensitivity analyses.} 
We conduct sensitivity analyses on DMC~(\textit{Cheetah Run} $\rightarrow$ \textit{Walker Walk}). 
As shown in \figref{fig:ablation_sensitivity_results} (Middle), we observe that when $\beta$ for the representation disentanglement is too small, the model learns entangled latent representations.
When $\beta$ is too large, it will impede the reconstruction of the image, leading to a decline in performance.
Latent distillation weight $\eta$ controls the cross-domain transfer scale. Intuitively, setting this hyperparameter too low may result in the downstream agent not getting enough knowledge from the pretrained models.
Conversely, excessively high $\eta$ may result in the model overfitting to the pretrained model, which is not conducive to the learning of the downstream task.
Additional sensitivity analyses on the latent space dimension are provided in \appref{sec:sens_latent_dim}.

\begin{figure}[htb]
    \centering
    \includegraphics[width=\linewidth]{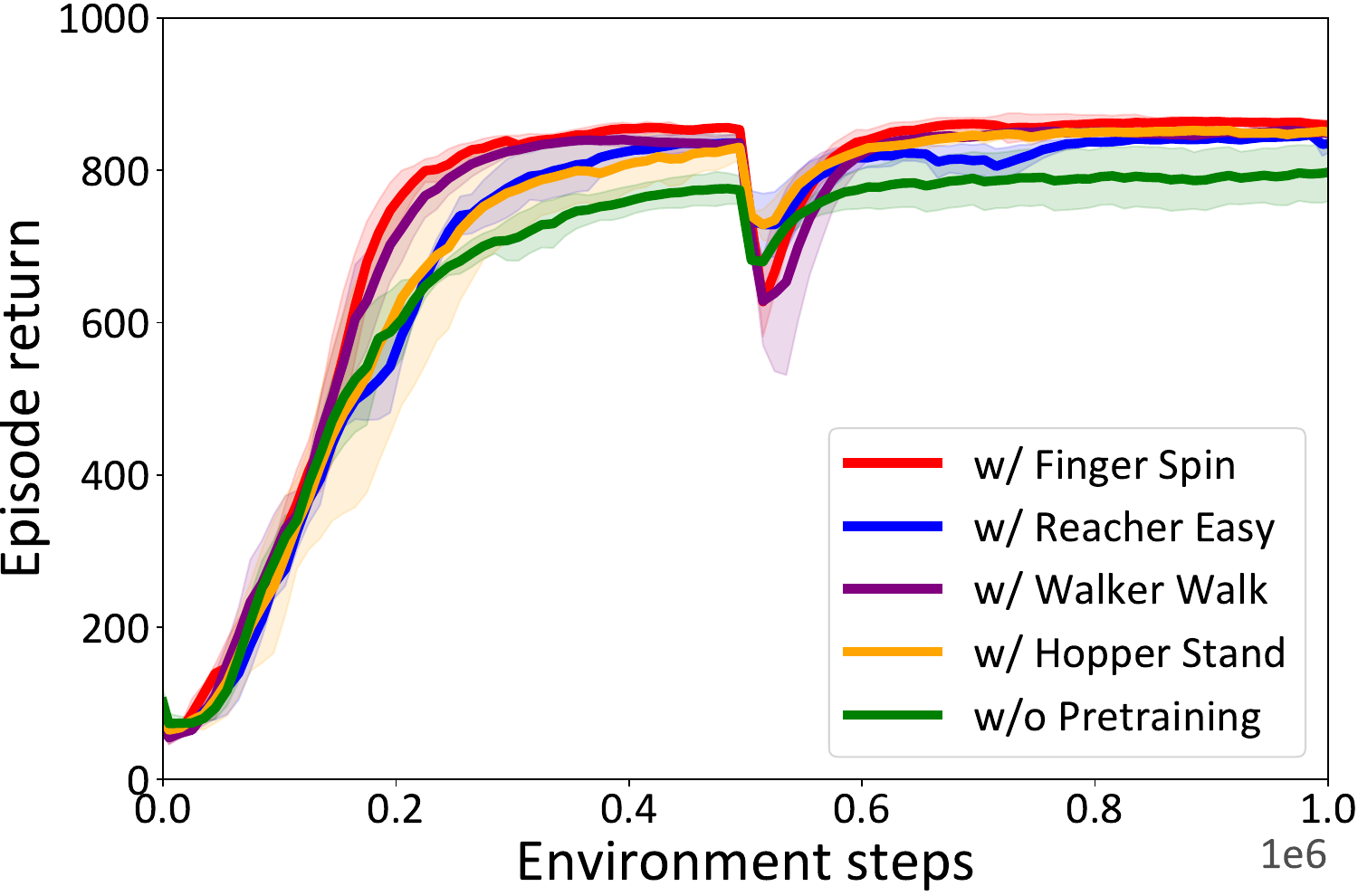}
    \caption{Performance of \model{} on DMC \ti{Cartpole Swingup}
    with different video datasets.}
    \label{fig:domain_selection}
\end{figure}

\paragraph{Effects of video domain.}
In \figref{fig:domain_selection}, we evaluate \model{} on DMC \ti{Cartpole Swingup} by pretraining on alternative video datasets, including frames collected from \ti{Finger Spin}, \ti{Reacher Easy}, \ti{Walker Walk},  and \ti{Hopper Stand}. 
Interestingly, compared with the DreamerV2 agent without pretraining, \model{} can always benefit from pretraining via offline-to-online latent distillation. 
It obtains the semantic knowledge from the pretrained model and strengthens the disentanglement capability during finetuning phase.

\label{sec:exp}


\section{Related Work}
\paragraph{Visual MBRL.}
Visual RL learns control policies from raw pixels, which has achieved remarkable performance in various tasks~\cite{wang2021unsupervised,choi2023environment,choi2023local,song2024simple}, while prior RL studies focus on learning policies from low-dimensional states. 
Extant approaches can be divided into two main directions: model-free RL~\citep{laskin2020curl,zhang2021dbc,stooke2021decoupling,yarats2022mastering,ma2023vip,zheng2024taco,li2024generalizing} and model-based RL~\citep{hafner2019learning,hafner2020dream,hafner2021mastering,hafner2025dreamerv3,zhang2023storm,seo2023multi,ma2024harmonydream,lin2024learning,alonso2024diffusion,mazzaglia2024genrl,wang2024making,li2025open}.
The following methods specifically address the variations of environments in visual MBRL.
Pan~\citep{pan2022iso} \etal~decompose visual dynamics into controllable and uncontrollable states through the optimization of inverse dynamics.
SeeX~\cite{huang2024leveraging} proposed a bi-level optimization framework that adopts a separated world model and maximizes the task-relevant uncertainty.
Orthogonal to these studies, our approach employs a DRL-based world model to alleviate the issue of visual variations.

\paragraph{Transfer RL}
To facilitate the learning of unseen tasks, transfer RL~\cite{sun2022transfer,zhu2023transfer,lu2023structured,ma2023vip,mazzaglia2023choreographer,wang2024making} leverages the knowledge learned from past tasks.
One promising way is to transfer world knowledge from accessible videos to improve the downstream control.
APV \cite{seo2022reinforcement} established a pretraining-finetuning framework with a stacked latent prediction model and video-based intrinsic bonus.
%
IPV \cite{wu2023pre} introduced contextualized world models that pretrained on diverse in-the-wild videos. It incorporates a context encoder that works alongside the latent dynamics model into the image encoder to capture rich contextual information.  
PreLAR~\cite{zhang2024prelar} pretrained the world model with the derived meaningful actions from the action-free video using an inverse dynamics encoder.
Different from these approaches, we propose a new solution to transfer world knowledge from distracting videos to improve the learning efficiency of downstream tasks via offline-to-online latent distillation.

\section{Conclusions and Limitations}
\label{sec:conclusion}
In this paper, we present a transfer RL dubbed \model{}, which addresses the challenge of environment variations in practical scenarios.
Our key insight is to leverage the accessible distracting videos to facilitate the sample efficiency of downstream tasks to offer flexible disentanglement constraints.
Specifically, we introduce \textit{disentangled representation pretraining}, \textit{offline-to-online latent distillation}, and \textit{disentangled world model adaptation} to improve the downstream control.
\model{} demonstrates superior performance than existing visual RL baselines across various benchmarks.

One limitation of our approach is that disentangled representation learning encounters challenges in complex environments. Exploring the non-stationary environments with more intricate variations, such as time-varying background video distractions, could further highlight the potential of our approach for practical scenarios.

\paragraph{Acknowledgements.} 
This work was supported by Grants of NSFC 62302246 \& 62250062, ZJNSFC LQ23F010008, Ningbo 2023Z237 \& 2024Z284 \& 2024Z289 \& 2023CX050011 \& 2025Z038, the Smart Grid National Science and Technology Major Project (2024ZD0801200), the Shanghai Municipal Science and Technology Major Project (2021SHZDZX0102), the Fundamental Research Funds for the Central Universities, and the IDT Foundation of Youth Doctoral Innovation~(S203.2.01.32.002).
Additional support was provided by the project of Supporting Program for Young and Middle-aged Scientific and Technological Innovation Talents in Shenyang (Grant RC210488), the project of Provincial Doctoral Research Initiation Fund Program (Grant 2023-BS-214), the High Performance Computing Center at Eastern Institute of Technology, Ningbo, and Ningbo Institute of Digital Twin.

{
    \small


\begin{thebibliography}{48}
\providecommand{\natexlab}[1]{#1}
\providecommand{\url}[1]{\texttt{#1}}
\expandafter\ifx\csname urlstyle\endcsname\relax
  \providecommand{\doi}[1]{doi: #1}\else
  \providecommand{\doi}{doi: \begingroup \urlstyle{rm}\Url}\fi

\bibitem[Alonso et~al.(2024)Alonso, Jelley, Micheli, Kanervisto, Storkey, Pearce, and Fleuret]{alonso2024diffusion}
Eloi Alonso, Adam Jelley, Vincent Micheli, Anssi Kanervisto, Amos Storkey, Tim Pearce, and Fran{\c{c}}ois Fleuret.
\newblock Diffusion for world modeling: Visual details matter in atari.
\newblock In \emph{NeurIPS}, 2024.

\bibitem[Bengio et~al.(2013)Bengio, Courville, and Vincent]{bengio2013representation}
Yoshua Bengio, Aaron Courville, and Pascal Vincent.
\newblock Representation learning: A review and new perspectives.
\newblock \emph{TPAMI}, 35\penalty0 (8):\penalty0 1798--1828, 2013.

\bibitem[Choi et~al.(2023{\natexlab{a}})Choi, Lee, Jeong, and Min]{choi2023environment}
Hyesong Choi, Hunsang Lee, Seongwon Jeong, and Dongbo Min.
\newblock Environment agnostic representation for visual reinforcement learning.
\newblock In \emph{ICCV}, pages 263--273, 2023{\natexlab{a}}.

\bibitem[Choi et~al.(2023{\natexlab{b}})Choi, Lee, Song, Jeon, Sohn, and Min]{choi2023local}
Hyesong Choi, Hunsang Lee, Wonil Song, Sangryul Jeon, Kwanghoon Sohn, and Dongbo Min.
\newblock Local-guided global: Paired similarity representation for visual reinforcement learning.
\newblock In \emph{CVPR}, pages 15072--15082, 2023{\natexlab{b}}.

\bibitem[Clevert et~al.(2015)Clevert, Unterthiner, and Hochreiter]{clevert2015fast}
Djork-Arn{\'e} Clevert, Thomas Unterthiner, and Sepp Hochreiter.
\newblock Fast and accurate deep network learning by exponential linear units (elus).
\newblock \emph{arXiv preprint arXiv:1511.07289}, 2015.

\bibitem[Dunion et~al.(2023{\natexlab{a}})Dunion, McInroe, Luck, Hanna, and Albrecht]{dunion2023conditional}
Mhairi Dunion, Trevor McInroe, Kevin~Sebastian Luck, Josiah Hanna, and Stefano Albrecht.
\newblock Conditional mutual information for disentangled representations in reinforcement learning.
\newblock In \emph{NeurIPS}, 2023{\natexlab{a}}.

\bibitem[Dunion et~al.(2023{\natexlab{b}})Dunion, McInroe, Luck, Hanna, and Albrecht]{dunion2023temporal}
Mhairi Dunion, Trevor McInroe, Kevin~Sebastian Luck, Josiah~P Hanna, and Stefano~V Albrecht.
\newblock Temporal disentanglement of representations for improved generalisation in reinforcement learning.
\newblock In \emph{ICLR}, 2023{\natexlab{b}}.

\bibitem[Hafner et~al.(2019)Hafner, Lillicrap, Fischer, Villegas, Ha, Lee, and Davidson]{hafner2019learning}
Danijar Hafner, Timothy Lillicrap, Ian Fischer, Ruben Villegas, David Ha, Honglak Lee, and James Davidson.
\newblock Learning latent dynamics for planning from pixels.
\newblock In \emph{ICML}, 2019.

\bibitem[Hafner et~al.(2020)Hafner, Lillicrap, Ba, and Norouzi]{hafner2020dream}
Danijar Hafner, Timothy Lillicrap, Jimmy Ba, and Mohammad Norouzi.
\newblock Dream to control: Learning behaviors by latent imagination.
\newblock In \emph{ICLR}, 2020.

\bibitem[Hafner et~al.(2021)Hafner, Lillicrap, Norouzi, and Ba]{hafner2021mastering}
Danijar Hafner, Timothy Lillicrap, Mohammad Norouzi, and Jimmy Ba.
\newblock Mastering atari with discrete world models.
\newblock In \emph{ICLR}, 2021.

\bibitem[Hafner et~al.(2022)Hafner, Lee, Fischer, and Abbeel]{hafner2022deep}
Danijar Hafner, Kuang-Huei Lee, Ian Fischer, and Pieter Abbeel.
\newblock Deep hierarchical planning from pixels.
\newblock \emph{arXiv preprint arXiv:2206.04114}, 2022.

\bibitem[Hafner et~al.(2025)Hafner, Pasukonis, Ba, and Lillicrap]{hafner2025dreamerv3}
Danijar Hafner, Jurgis Pasukonis, Jimmy Ba, and Timothy Lillicrap.
\newblock Mastering diverse domains through world models.
\newblock \emph{Nature}, 2025.

\bibitem[Hansen et~al.(2024)Hansen, Su, and Wang]{hansen2024td}
Nicklas Hansen, Hao Su, and Xiaolong Wang.
\newblock Td-mpc2: Scalable, robust world models for continuous control.
\newblock In \emph{ICLR}, 2024.

\bibitem[Higgins et~al.(2017{\natexlab{a}})Higgins, Matthey, Pal, Burgess, Glorot, Botvinick, Mohamed, and Lerchner]{higgins2017beta}
Irina Higgins, Loic Matthey, Arka Pal, Christopher Burgess, Xavier Glorot, Matthew Botvinick, Shakir Mohamed, and Alexander Lerchner.
\newblock beta-vae: Learning basic visual concepts with a constrained variational framework.
\newblock In \emph{ICLR}, 2017{\natexlab{a}}.

\bibitem[Higgins et~al.(2017{\natexlab{b}})Higgins, Pal, Rusu, Matthey, Burgess, Pritzel, Botvinick, Blundell, and Lerchner]{higgins2017darla}
Irina Higgins, Arka Pal, Andrei Rusu, Loic Matthey, Christopher Burgess, Alexander Pritzel, Matthew Botvinick, Charles Blundell, and Alexander Lerchner.
\newblock Darla: Improving zero-shot transfer in reinforcement learning.
\newblock In \emph{ICML}, 2017{\natexlab{b}}.

\bibitem[Huang et~al.(2024)Huang, Wan, Shao, Sun, Gan, Feng, and Zhan]{huang2024leveraging}
Kaichen Huang, Shenghua Wan, Minghao Shao, Hai-Hang Sun, Le Gan, Shuai Feng, and De-Chuan Zhan.
\newblock Leveraging separated world model for exploration in visually distracted environments.
\newblock \emph{NeurIPS}, 2024.

\bibitem[Laskin et~al.(2020{\natexlab{a}})Laskin, Lee, Stooke, Pinto, Abbeel, and Srinivas]{laskin2020reinforcement}
Misha Laskin, Kimin Lee, Adam Stooke, Lerrel Pinto, Pieter Abbeel, and Aravind Srinivas.
\newblock Reinforcement learning with augmented data.
\newblock In \emph{NeurIPS}, 2020{\natexlab{a}}.

\bibitem[Laskin et~al.(2020{\natexlab{b}})Laskin, Srinivas, and Abbeel]{laskin2020curl}
Michael Laskin, Aravind Srinivas, and Pieter Abbeel.
\newblock Curl: Contrastive unsupervised representations for reinforcement learning.
\newblock In \emph{ICML}, pages 5639--5650, 2020{\natexlab{b}}.

\bibitem[Li et~al.(2024)Li, Jiang, CHEN, and Zhao]{li2024generalizing}
Haoran Li, Zhennan Jiang, YUHUI CHEN, and Dongbin Zhao.
\newblock Generalizing consistency policy to visual rl with prioritized proximal experience regularization.
\newblock In \emph{NeurIPS}, 2024.

\bibitem[Li et~al.(2025)Li, Wang, Wang, Jin, Li, Zeng, and Yang]{li2025open}
Jiajian Li, Qi Wang, Yunbo Wang, Xin Jin, Yang Li, Wenjun Zeng, and Xiaokang Yang.
\newblock Open-world reinforcement learning over long short-term imagination.
\newblock In \emph{ICLR}, 2025.

\bibitem[Lin et~al.(2024)Lin, Du, Watkins, Hafner, Abbeel, Klein, and Dragan]{lin2024learning}
Jessy Lin, Yuqing Du, Olivia Watkins, Danijar Hafner, Pieter Abbeel, Dan Klein, and Anca Dragan.
\newblock Learning to model the world with language.
\newblock In \emph{ICML}, 2024.

\bibitem[Lu et~al.(2023)Lu, Schroecker, Gu, Parisotto, Foerster, Singh, and Behbahani]{lu2023structured}
Chris Lu, Yannick Schroecker, Albert Gu, Emilio Parisotto, Jakob Foerster, Satinder Singh, and Feryal Behbahani.
\newblock Structured state space models for in-context reinforcement learning.
\newblock \emph{NeurIPS}, 2023.

\bibitem[Ma et~al.(2024)Ma, Wu, Feng, Xiao, Li, Hao, Wang, and Long]{ma2024harmonydream}
Haoyu Ma, Jialong Wu, Ningya Feng, Chenjun Xiao, Dong Li, Jianye Hao, Jianmin Wang, and Mingsheng Long.
\newblock Harmonydream: Task harmonization inside world models.
\newblock In \emph{ICML}, 2024.

\bibitem[Ma et~al.(2023)Ma, Sodhani, Jayaraman, Bastani, Kumar, and Zhang]{ma2023vip}
Yecheng~Jason Ma, Shagun Sodhani, Dinesh Jayaraman, Osbert Bastani, Vikash Kumar, and Amy Zhang.
\newblock Vip: Towards universal visual reward and representation via value-implicit pre-training.
\newblock In \emph{ICLR}, 2023.

\bibitem[Maaten and Hinton(2008)]{maaten2008visualizing}
Laurens van~der Maaten and Geoffrey Hinton.
\newblock Visualizing data using t-sne.
\newblock \emph{JMLR}, 9\penalty0 (Nov):\penalty0 2579--2605, 2008.

\bibitem[Mazzaglia et~al.(2023)Mazzaglia, Verbelen, Dhoedt, Lacoste, and Rajeswar]{mazzaglia2023choreographer}
Pietro Mazzaglia, Tim Verbelen, Bart Dhoedt, Alexandre Lacoste, and Sai Rajeswar.
\newblock Choreographer: Learning and adapting skills in imagination.
\newblock In \emph{ICLR}, 2023.

\bibitem[Mazzaglia et~al.(2024)Mazzaglia, Verbelen, Dhoedt, Courville, and Mudumba]{mazzaglia2024genrl}
Pietro Mazzaglia, Tim Verbelen, Bart Dhoedt, Aaron~C Courville, and Sai~Rajeswar Mudumba.
\newblock Genrl: Multimodal-foundation world models for generalization in embodied agents.
\newblock \emph{NeurIPS}, 2024.

\bibitem[Pan et~al.(2022)Pan, Zhu, Wang, and Yang]{pan2022iso}
Minting Pan, Xiangming Zhu, Yunbo Wang, and Xiaokang Yang.
\newblock Iso-dream: Isolating and leveraging noncontrollable visual dynamics in world models.
\newblock In \emph{NeurIPS}, 2022.

\bibitem[Seo et~al.(2022)Seo, Lee, James, and Abbeel]{seo2022reinforcement}
Younggyo Seo, Kimin Lee, Stephen~L James, and Pieter Abbeel.
\newblock Reinforcement learning with action-free pre-training from videos.
\newblock In \emph{ICML}, 2022.

\bibitem[Seo et~al.(2023)Seo, Kim, James, Lee, Shin, and Abbeel]{seo2023multi}
Younggyo Seo, Junsu Kim, Stephen James, Kimin Lee, Jinwoo Shin, and Pieter Abbeel.
\newblock Multi-view masked world models for visual robotic manipulation.
\newblock In \emph{ICML}, 2023.

\bibitem[Song et~al.(2024)Song, Choi, Sohn, and Min]{song2024simple}
Wonil Song, Hyesong Choi, Kwanghoon Sohn, and Dongbo Min.
\newblock A simple framework for generalization in visual rl under dynamic scene perturbations.
\newblock \emph{NeurIPS}, 37:\penalty0 121790--121826, 2024.

\bibitem[Stooke et~al.(2021)Stooke, Lee, Abbeel, and Laskin]{stooke2021decoupling}
Adam Stooke, Kimin Lee, Pieter Abbeel, and Michael Laskin.
\newblock Decoupling representation learning from reinforcement learning.
\newblock In \emph{ICML}, pages 9870--9879, 2021.

\bibitem[Sun et~al.(2022)Sun, Zheng, Wang, Cohen, and Huang]{sun2022transfer}
Yanchao Sun, Ruijie Zheng, Xiyao Wang, Andrew Cohen, and Furong Huang.
\newblock Transfer rl across observation feature spaces via model-based regularization.
\newblock In \emph{ICLR}, 2022.

\bibitem[Tassa et~al.(2018)Tassa, Doron, Muldal, Erez, Li, Casas, Budden, Abdolmaleki, Merel, Lefrancq, et~al.]{tassa2018deepmind}
Yuval Tassa, Yotam Doron, Alistair Muldal, Tom Erez, Yazhe Li, Diego de~Las Casas, David Budden, Abbas Abdolmaleki, Josh Merel, Andrew Lefrancq, et~al.
\newblock Deepmind control suite.
\newblock \emph{arXiv preprint arXiv:1801.00690}, 2018.

\bibitem[Todorov et~al.(2012)Todorov, Erez, and Tassa]{todorov2012mujoco}
Emanuel Todorov, Tom Erez, and Yuval Tassa.
\newblock Mujoco: A physics engine for model-based control.
\newblock In \emph{IROS}, 2012.

\bibitem[Wang et~al.(2024{\natexlab{a}})Wang, Yang, Wang, Jin, Zeng, and Yang]{wang2024making}
Qi Wang, Junming Yang, Yunbo Wang, Xin Jin, Wenjun Zeng, and Xiaokang Yang.
\newblock Making offline rl online: Collaborative world models for offline visual reinforcement learning.
\newblock In \emph{NeurIPS}, 2024{\natexlab{a}}.

\bibitem[Wang et~al.(2021)Wang, Lian, and Yu]{wang2021unsupervised}
Xudong Wang, Long Lian, and Stella~X Yu.
\newblock Unsupervised visual attention and invariance for reinforcement learning.
\newblock In \emph{CVPR}, pages 6677--6687, 2021.

\bibitem[Wang et~al.(2024{\natexlab{b}})Wang, Chen, Wu, Zhu, et~al.]{wang2024disentangled}
Xin Wang, Hong Chen, Zihao Wu, Wenwu Zhu, et~al.
\newblock Disentangled representation learning.
\newblock \emph{TPAMI}, 2024{\natexlab{b}}.

\bibitem[Wu et~al.(2023)Wu, Ma, Deng, and Long]{wu2023pre}
Jialong Wu, Haoyu Ma, Chaoyi Deng, and Mingsheng Long.
\newblock Pre-training contextualized world models with in-the-wild videos for reinforcement learning.
\newblock \emph{NeurIPS}, 2023.

\bibitem[Xie et~al.(2023)Xie, Li, Zhang, Dong, Jin, Yang, and Zeng]{xie2023navinerf}
Baao Xie, Bohan Li, Zequn Zhang, Junting Dong, Xin Jin, Jingyu Yang, and Wenjun Zeng.
\newblock Navinerf: Nerf-based 3d representation disentanglement by latent semantic navigation.
\newblock In \emph{ICCV}, 2023.

\bibitem[Xie et~al.(2024)Xie, Chen, Wang, Zhang, Jin, and Zeng]{xie2024graph}
Baao Xie, Qiuyu Chen, Yunnan Wang, Zequn Zhang, Xin Jin, and Wenjun Zeng.
\newblock Graph-based unsupervised disentangled representation learning via multimodal large language models.
\newblock In \emph{NeurIPS}, 2024.

\bibitem[Yarats et~al.(2022)Yarats, Fergus, Lazaric, and Pinto]{yarats2022mastering}
Denis Yarats, Rob Fergus, Alessandro Lazaric, and Lerrel Pinto.
\newblock Mastering visual continuous control: Improved data-augmented reinforcement learning.
\newblock In \emph{ICLR}, 2022.

\bibitem[Yu et~al.(2019)Yu, Quillen, He, Julian, Hausman, Finn, and Levine]{yu2019meta}
Tianhe Yu, Deirdre Quillen, Zhanpeng He, Ryan Julian, Karol Hausman, Chelsea Finn, and Sergey Levine.
\newblock Meta-world: A benchmark and evaluation for multi-task and meta reinforcement learning.
\newblock In \emph{CoRL}, 2019.

\bibitem[Zhang et~al.(2021)Zhang, McAllister, Calandra, Gal, and Levine]{zhang2021dbc}
Amy Zhang, Rowan McAllister, Roberto Calandra, Yarin Gal, and Sergey Levine.
\newblock Learning invariant representations for reinforcement learning without reconstruction.
\newblock In \emph{ICLR}, 2021.

\bibitem[Zhang et~al.(2024)Zhang, Kan, Shan, and Chen]{zhang2024prelar}
Lixuan Zhang, Meina Kan, Shiguang Shan, and Xilin Chen.
\newblock Prelar: World model pre-training with learnable action representation.
\newblock In \emph{ECCV}, 2024.

\bibitem[Zhang et~al.(2023)Zhang, Wang, Sun, Yuan, and Huang]{zhang2023storm}
Weipu Zhang, Gang Wang, Jian Sun, Yetian Yuan, and Gao Huang.
\newblock Storm: Efficient stochastic transformer based world models for reinforcement learning.
\newblock In \emph{NeurIPS}, 2023.

\bibitem[Zheng et~al.(2023)Zheng, Wang, Sun, Ma, Zhao, Xu, Daum{\'e}~III, and Huang]{zheng2024taco}
Ruijie Zheng, Xiyao Wang, Yanchao Sun, Shuang Ma, Jieyu Zhao, Huazhe Xu, Hal Daum{\'e}~III, and Furong Huang.
\newblock Taco: Temporal latent action-driven contrastive loss for visual reinforcement learning.
\newblock In \emph{NeurIPS}, 2023.

\bibitem[Zhu et~al.(2023)Zhu, Lin, Jain, and Zhou]{zhu2023transfer}
Zhuangdi Zhu, Kaixiang Lin, Anil~K Jain, and Jiayu Zhou.
\newblock Transfer learning in deep reinforcement learning: A survey.
\newblock \emph{TPAMI}, 45\penalty0 (11):\penalty0 13344--13362, 2023.

\end{thebibliography}
}

\clearpage
\setcounter{page}{1}
\setcounter{section}{0}
\setcounter{table}{0}
\maketitlesupplementary

\appendix

\renewcommand\thesection{\Alph{section}}
\renewcommand\thetable{\Alph{table}}
\renewcommand\thefigure{\Alph{figure}}

\section{Compared Baselines}
\label{sec:compared_baselines}
We compare \model{} with strong visual RL agents, including
\begin{itemize}[leftmargin=*]
    \item \textbf{DreamerV2}~\citep{hafner2021mastering}:
    A model-based RL~(MBRL) approach that trains world model and learns by imagining future latent states.
    \item \textbf{APV}~\citep{seo2022reinforcement}: It learns informational representations via action-free pretraining on videos and finetunes the agent with learned representations in the downstream tasks with action.
    \item \textbf{DV2 Finetune}: It pretrains a DreamerV2 agent~\cite{hafner2021mastering} on distracting videos and then finetunes the trained model in the downstream tasks.
    Note that some tasks have different action spaces, which makes it difficult to finetune directly. Therefore, the action space of two tasks is set as the maximum action space of both environments.
    \item \textbf{TED}~\citep{dunion2023temporal}: It adopts a classification task to learn temporally disentangled representations in visual RL.
    \item \textbf{CURL}~\citep{laskin2020curl}: 
    A model-free RL method that employs contrastive learning to improve its sample efficiency.
\end{itemize}

\section{Behavior Learning}
\label{sec:bl}
For the behavior learning of \model{}, we adopt the actor-critic method following DreamerV2~\citep{hafner2021mastering}.
Concretely, the actor and critic are both implemented as MLPs with ELU activations~\cite{clevert2015fast}. Formally, the actor and critic are defined as below:
\begin{equation}
\begin{alignedat}{3}
& \text{Actor:}~    && \hat{a}_t \sim \pi_{\psi}(\hat{a}_t | \hat{z}_t) \\
& \text{Critic:}~   && v_{\xi}(\hat{z}_t) \approx \mathbb{E}_{p_{\phi},p_{\psi}}\Big[
  \textstyle\sum_{\tau \geq t} \hat{\gamma}_{\tau-t} \hat{r}_\tau
\Big]. \\
\end{alignedat}
\end{equation}
The  actor $\pi_{\psi}$ is optimized by maximizing 
\begin{equation}
\label{eq:actor_loss}
\begin{aligned}
    &\mathcal{L}(\psi)  = \ \mathbb{E}_{p_{\phi}, p_{\psi}} \Big[\sum_{t=1}^{H-1}(
    \underbrace{\beta \mathrm{H} \left[a_t \mid \hat{z}_t \right]}_{\text{entropy regularization}}
    + \underbrace{\rho V_t}_{\text{dynamics backprop}} \\ &+ 
    \underbrace{(1-\rho) \ln \pi_{\psi}(\hat{a}_t \mid \hat{z}_t) \texttt{sg}(V_t-v_{\xi}(\hat{z}_t))}_{\text{REINFORCE}}\Big].
\end{aligned}
\end{equation}
We train the critic $v_{\xi}$ by minimizing
\begin{equation}
\label{eq:critic_loss}
\begin{aligned}
    \mathcal{L}(\xi) = \mathbb{E}_{p_{\phi}, p_{\psi}}\Big[\sum_{t=1}^{H-1} \frac{1}{2}\left(v_{\xi}\left(\hat{z}_t\right)-\texttt{sg}\left(V_t\right)\right)^2\Big].
\end{aligned}
\end{equation}
where $\texttt{sg}$ is a stop gradient operator.

The $\lambda$-target $V_t$ that involves a weighted average of reward information used in \eqref{eq:actor_loss} and \eqref{eq:critic_loss} is defined as:
\begin{equation}
V_t \doteq \hat{r}_t+\hat{\gamma}_t \begin{cases}(1-\lambda) v_{\xi}\left(\hat{z}_{t+1}\right)+\lambda V_{t+1} & \text { if } t<H \\ v_{\xi}\left(\hat{z}_H\right) & \text { if } t=H\end{cases}.
\end{equation}
where $H$ is the imagination horizon.
Notably, the disentangled world model is \ti{not} optimized during behavior learning.

\section{Additional Results}
\subsection{Results on DMC}
\label{sec:res_dmc}
We compare the performance of \ti{DreamerV3}~\cite{hafner2025dreamerv3}, \ti{TD-MPC2}~\cite{hansen2024td}, ContextWM~\cite{wu2023pre}, and our approach on DMC. As shown \bb{\tabref{tab:compare_baselines}}, \model{} outperforms other strong baselines in terms of episode return. 

\begin{table*}[th]
\setlength{\tabcolsep}{3.pt}
\centering
\begin{center}
\begin{tabular}{l | c c c}
\hline
Model & \ti{Reacher Easy} $\rightarrow$ \ti{Cheetah Run} &\ti{Walker Walk} $\rightarrow$ \ti{Humanoid Walk}  \\
\hline
DreamerV3 & 662 $\pm$ 9 & 12 $\pm$ 17 \\
TD-MPC2 & 510 $\pm$ 15 & 1 $\pm$ 0 \\
ContextWM & 661 $\pm$ 49 & 1 $\pm$ 0 \\
DisWM & \bb{817 $\pm$ 59} & \bb{147 $\pm$ 85} \\
\hline
\end{tabular}
\caption{Comparison with strong baselines on DMC.}
\label{tab:compare_baselines}
\end{center}
\end{table*}

\subsection{Results on DrawerWorld}
\label{sec:results_drawerworld}
We present results on DrawerWorld~\cite{wang2021unsupervised} in \bb{\tabref{tab:drawerworld_res}}. 
As reported in \bb{\tabref{tab:drawerworld_res}}, DisWM~(source: \ti{Finger Spin}) outperforms other baselines in terms of success rate (\%) on all tasks.

\begin{table}[th]
  \setlength{\tabcolsep}{5.pt}
  \centering
  \begin{center}
  \vspace{-3.mm}
  \begin{tabular}{l | c c c}
  \hline
  
  Model & DrawerClose & DrawerOpen \\
  \hline
TDMPC2      &  3 $\pm$ 6   & 43 $\pm$ 25 \\
ContextWM   & 37 $\pm$ 12   & 23 $\pm$ 25 \\
DisWM       & \bb{77 $\pm$ 6} & \bb{70 $\pm$ 10}  \\
  \hline
  \end{tabular}
  \caption{Performance on DrawerWorld with texture variations.}
  \label{tab:drawerworld_res}
  \end{center}
\end{table}

\subsection{Sensitivity of the Latent Space Dimension}
\label{sec:sens_latent_dim}
\begin{figure}[htb]
    \centering
\includegraphics[width=\linewidth]{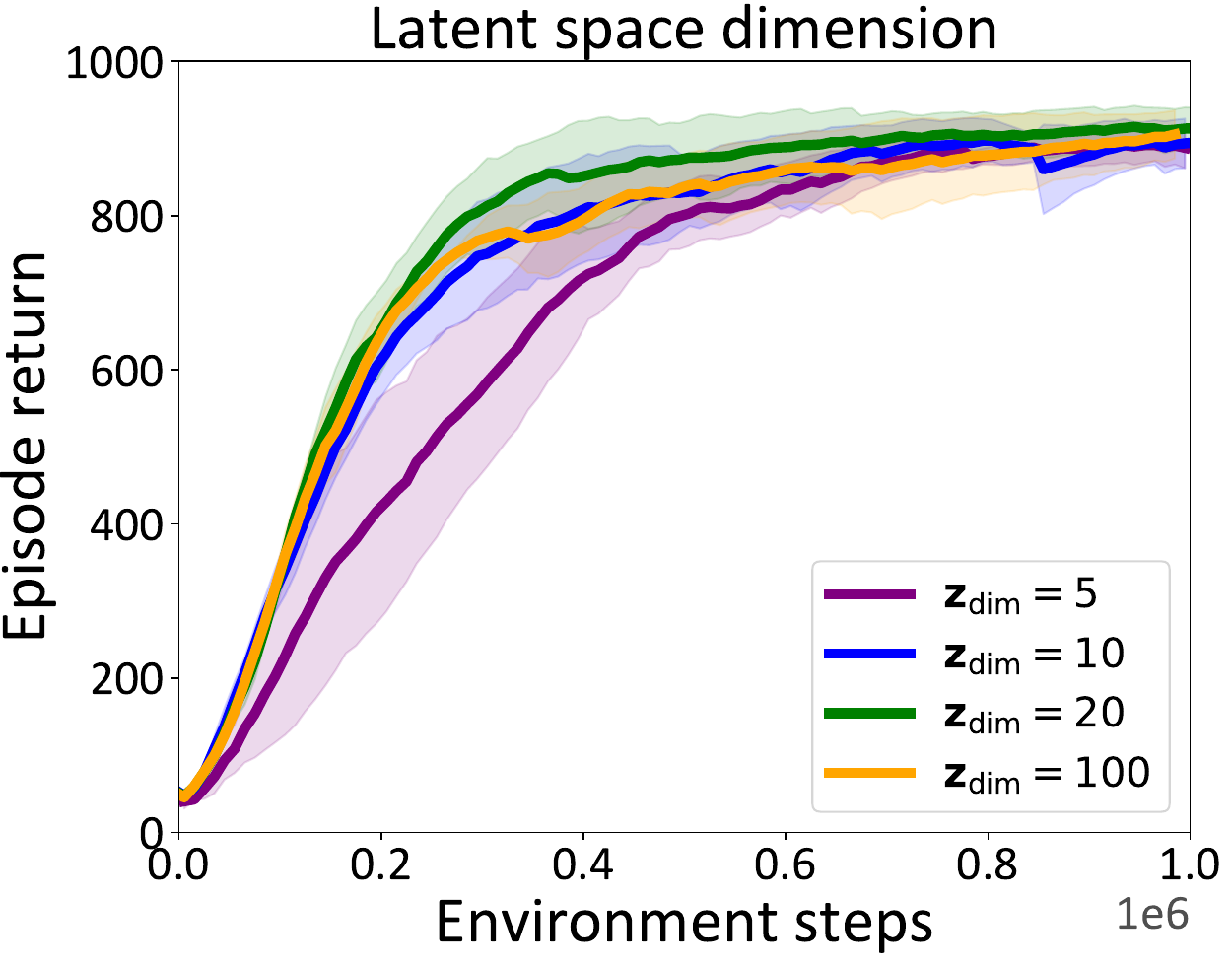}
    \caption{Sensitivity analyses on \ti{Cheetah Run} $\rightarrow$ \ti{Walker Walk}}
    \label{fig:dimension_compare}
\end{figure}

We visualize sensitivity analyses on the latent space dimension in \figref{fig:dimension_compare}. We observe that when $\mathbf{z}_\text{dim}$ for the $\beta$-VAE is too small, it impedes the learning of disentangled representations, leading to a decline in performance.

\subsection{Runtime Comparisons}
We provide the detailed runtime and parameter comparisons with baselines in \tabref{tab:time_complexity}. Note that the inference time is computed for one episode.

\begin{table}[th]
\footnotesize
\caption{Runtime and model size comparisons evaluated on DMC (\textit{Finger Spin} $\rightarrow$ \textit{Reacher Easy}). DV2 FT is short for DreamerV2 finetune.} 
\vspace{-10pt}
\label{tab:time_complexity}
\setlength\tabcolsep{1.2mm}
\begin{center}
\begin{tabular}{l|cccc}
\toprule

Model& Training Steps & Training time&  Inference time& Params (M) \\
\midrule

CURL & 100k & 303 min &4.97 sec & 10.7 \\
DV2 FT & 200k & 1522 min &9.88 sec & 12.1 \\
APV & 200k & 1722 min &10.15 sec & 13 \\
TED & 100k & 1051 min &20.49 sec & 11.5 \\
DV2 & 100k & 901 min &9.59 sec & 12.1 \\
\model{} & 200k & 1311 min &9.48 sec & 5.8 \\

\bottomrule
\end{tabular}

\end{center}
\end{table}

\subsection{Sample Diversity Visualization}
\begin{figure}[htb]
    \centering    \includegraphics[width=0.9\linewidth]{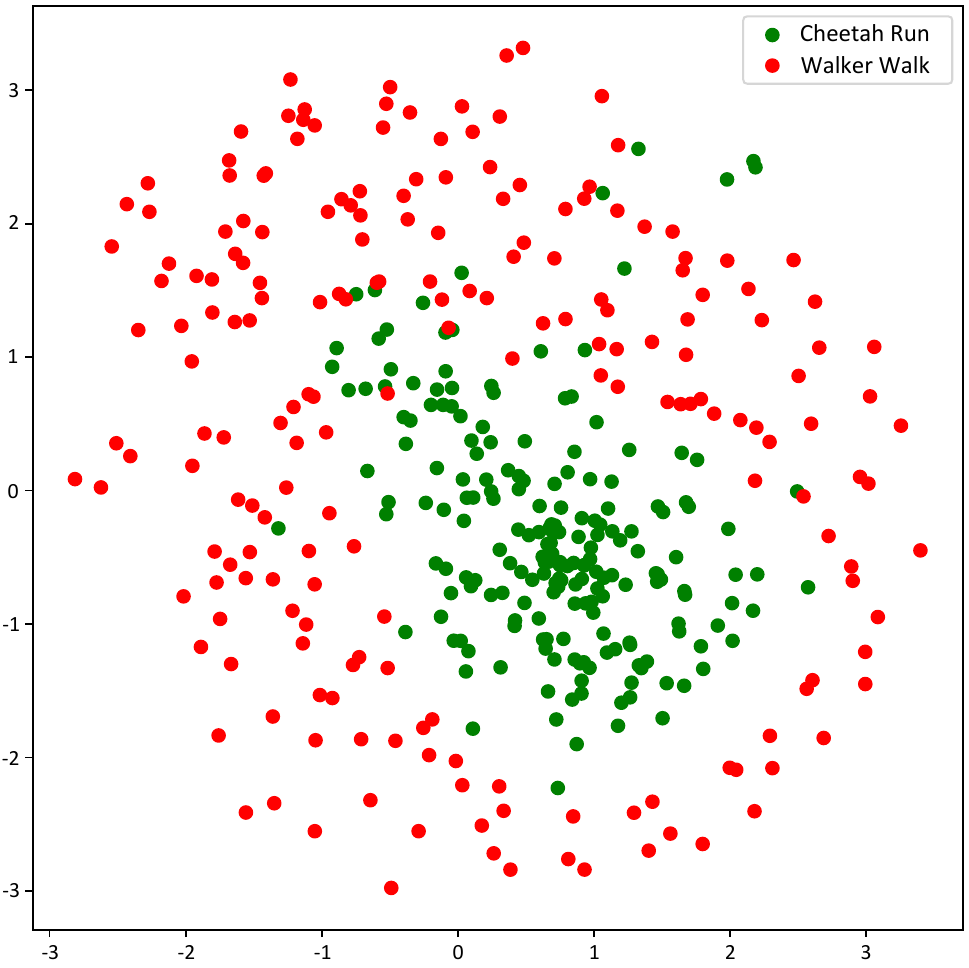}
    \caption{Sample diversity enhanced by adaptation.}
    \label{fig:distribution}
\end{figure}

The adaptation stage enriches the sample diversity, as shown in \figref{fig:distribution}, for \ti{Cheetah Run} $\rightarrow$ \ti{Walker Walk}, we sample 200 video clips of length 50 and visualize the corresponding latent features using t-SNE~\cite{maaten2008visualizing}. 
We find that the latent features of the online interactions are more diverse than those of the offline dataset.

\section{Hyperparameters}

The final hyperparameters of \model{} are reported in \tabref{tab:hparams}. 
\begin{table}[H]
\centering
\caption{Hyperparameters of \model{}.} 
\vspace{-3pt}
\vskip 0.05in
\setlength{\tabcolsep}{0.6mm}{} %
\begin{tabular}{lccc}
\toprule
\textbf{Name} & \textbf{Notation} & \textbf{Value} \\
\midrule
\texttt{Video prediction model}  \\
\midrule
Image size & --- & $64 \times 64$ \\
KL divergence scale & $\beta_1$ & $1$ \\
Disentanglement scale & $\beta_2$ & $0.015$ \\
Latent dimension & --- & $20$ \\
Learning rate & --- & $3\cdot10^{-4}$ \\
\midrule
\texttt{Disentangled World Model} \\
\midrule
Latent distillation weight & $\eta$ & $0.1$ \\
Disentanglement scale & $\beta$ & $0.015$ \\
KL divergence scale & $\alpha$ & $1$ \\
Latent dimension & --- & $20$ \\
Batch size & $B$ & $50$ \\
Batch length & $L$ & $50$ \\
Learning rate & --- & $3\cdot10^{-4}$ \\

\midrule
\texttt{Behavior Learning} \\
\midrule
Imagination horizon & $H$ & $15$ \\
Discount & $\gamma$ & $0.99$ \\
$\lambda$-target  & $\lambda$ & $0.95$ \\
Actor learning rate & --- & $8\cdot10^{-5}$ \\
Critic learning rate & --- & $8\cdot10^{-5}$ \\
\bottomrule
\end{tabular}
\label{tab:hparams}
\end{table}

\end{document}